# Performance Evaluation of Sentiment Analysis on Text and Emoji Data Using End-to-End, Transfer Learning, Distributed and Explainable AI Models

Sirisha Velampalli[1], Chandrashekar Muniyappa[2], and Ashutosh Saxena[1]
[1] CR Rao AIMSCS, University of Hyderabad Campus, Hyderabad, India
[2] Independent Researcher, Dublin, U.S.A.
Email: {sirisha.crraoaimscs, cnachiketa07, saxenaaj}@gmail.com

*Abstract*—Emojis are being frequently used in today's digital world to express from simple to complex thoughts more than ever before. Hence, they are also being used in sentiment analysis and targeted marketing campaigns. In this work, we performed sentiment analysis of Tweets as well as on emoji dataset from the Kaggle. Since tweets are sentences we have used Universal Sentence Encoder (USE) and Sentence Bidirectional Encoder Representations from Transformers (SBERT) end-to-end sentence embedding models to generate the embeddings which are used to train the Standard fully connected Neural Networks (NN), and LSTM NN models. We observe the text classification accuracy was almost the same for both the models around 98%. On the contrary, when the validation set was built using emojis that were not present in the training set then the accuracy of both the models reduced drastically to 70%. In addition, the models were also trained using the distributed training approach instead of a traditional single-threaded model for better scalability. Using the distributed training approach, we were able to reduce the run-time by roughly 15% without compromising on accuracy. Finally, as part of explainable AI the Shap algorithm was used to explain the model behaviour and check for model biases for the given feature set.

*Index Terms*—emoji, embedding models, sentiment analysis, distributed machine learning, explainable artificial intelligence

## I. INTRODUCTION

Sentiment analysis [1] or opinion mining is a Natural Language Processing (NLP) technique that helps to determine whether the opinion is positive, negative, or neutral. Some of the applications include identifying online trends, analyzing reviews, monitor brand and product market based on sentiments in customer feedback. Sentiment analysis typically follows a general framework which includes collecting raw data, preprocessing the collected data to remove noise, transforming the preprocessed data to the computational suitable form, labeling the data for training. In addition, sentiment analysis algorithms come in Three main forms namely rule-based, automatic, and hybrid. Rule-based algorithms are based on manually crafted rules, whereas automatic algorithms rely on machine learning techniques and hybrid algorithms combine both rule-based and automatic approaches.

Nowadays in social media texts like blogs, micro-blogs (Ex: Twitter), chats (Ex: WhatsApp and Facebook) people are using emojis more than text [2]. For the first time ever, in November 2015 oxford dictionaries word of the year was chosen to be an Emoji character which is known as "Face with Tears of Joy" or the expressivity of a text message can be enhanced with a single emoji character. A smiling face Emoji character can express someone's positive feeling whereas an angry face Emoji may reveal negative feelings. One can get a clear sense of the emotion in the text based on emoji characters used in it. In this work, we performed sentiment analysis of Tweets that contains emojis as well as on the emoji dataset from Kaggle.

The number of users on social networks is increasing constantly. Users post information about various topics [3]. The information can help researchers, manufacturers, politicians, and celebrities in understanding the user or market sentiments and make required changes based on the insights gained to improve their popularity. Twitter is one of the most popular social networking sites. Twitter users are increasing day by day with approximately 500 million tweets published daily. Tweet sentiment analysis is increasingly applied in many areas, such as decision support systems and recommendation systems. Therefore, improving the accuracy of tweet sentiment analysis has become pivotal and an area of interest for many researchers. Many studies [4], [5] have tried to improve the performance of tweet sentiment analysis methods by using only textual data. In this work, the H4EAD data was collected for a week using the Twitter search API over a period of one week. The positive tweets were collected by querying the "saveh4ead" hashtag and the negative tweets were collected by querying the "noh4ead" hashtag. There were 2252 tweets, out of which 246 were negative and 2006 were positive tweets. We have also conducted experiments on the Kaggle dataset that contain positive and negative emojis and the results are quite encouraging.

The rest of the paper is organized as follows:

The problem statement and objectives are explained in Section II. Related work is discussed in Section III. Overall Methodology is presented in Section IV. Conducted Experiments, Results and Analysis are discussed in Sections V and Section VI respectively. Finally, the Conclusion and future scope are presented in Section VII.

## II. PROBLEM STATEMENT AND OBJECTIVES

Sentiment Analysis (SA) has many applications in the real world. Many researchers [1]-[3], [6] addressed the problem of SA from text, images, emoticon, audio or video separately. However, only a few researchers have worked on emojis for finding sentiments. Also to our knowledge, none applied distributed machine learning and Explainable AI (XAI) on emoji datasets. Objectives of our work are as follows:

- Sentiment analysis of Tweets as well as on emoji dataset from Kaggle.
- To generate embeddings using Universal Sentence Encoder and SBERT sentence embedding models.
- To improve the classification accuracy using Standard fully connected NN, and LSTM NN models.
- Train the models using distributed training approach instead of a traditional single-threaded model for better scalability.
- To explain the model behaviour and check for model biases based on the given featureset using explainable AI.

## III. RELATED WORK

In [6] authors have applied BOW (Bag-Of-Words) and TF-IDF which are mainly based on keyword targeting and do not capture the semantic relationship between different words having a similar meaning, as a result, it will have low accuracy. Automatic detection [7] of the Tympanic Membrane (TM) and Middle Ear (ME) infection is done using state-of-the-art CNN models such as DenseNet. In [8], authors explored the emoji modality challenges that arise through the lens of multimedia research. They collected a large-scale data set of emojis from Twitter. The Dataset contained both text and emojis, using the state-of-the-art neural networks authors were able to predict emojis. In [9] by taking the Twitter dataset, authors extended binary sentiment classification approaches using a multi-way emotions classification. In [4] authors pre-processed Twitter messages and they have used R and Rapid Miner tools to categorize the sentiments into neutral, negative, and positive.

In [10] authors experimented with different classification methods like Naive Bayes (NB), SVM, Decision Tree, and Random Forest (RF). They have taken a Twitter dataset that contains 12864 tweets with 10 fold validation. In [11] authors applied deep learning to identify the seven main human emotions: anger, disgust, fear, happiness, sadness, surprise, and neutrality. Sentiment analysis was implemented on the Kazakh language via Spark on the basis of data set taken from Kazakh books in [12]. In [13] authors performed sentiment analysis on educational big data using multi-attention fusion modeling which integrates global attention and local attention through gating unit control to generate a reasonable contextual representation. Unlike the existing approaches, to solve this problem, we have used Sentence embedding models like S-BERT and Universal Sentence Encoders (USE) which are designed specifically to handle semantic relationships and generate fixed length of embeddings without the need to manually pad the embeddings for sentences of various lengths as described in earlier works. In addition, as these are end-to-end models there is no need to manually clean the data. The effect of these improvements is evident in the improved LSTM results with 98% accuracy.

## IV. METHODOLOGY

Data for this research was collected from Twitter and Kaggle. The H4EAD data was collected using the tweeter search API over a period of one week. There were 2252 tweets, out of which 246 were negative and 2006 were positive tweets. We performed sentiment analysis of tweets as well as on emoji dataset from the Kaggle. Since tweets are sentences we have used sentence embedding models like Universal Sentence Embedding and SBERTS to embed sentences to sentence vectors instead of the basic bag of words (BOW) models like TF-IDF and Count vectors. We applied Standard Fully connected Neural Networks (NN) and LSTM NN models. For better scalability, we performed distributed training and finally as part of explainable AI the Shap algorithm was used to check for model behaviour and model biases for the given feature set. Overall methodology is shown in the Fig. 1. Proposed methodology is shown in the form of algorithmic steps in Algorithm 1.

---

Algorithm 1 *Data with and without emojis*

---

1: **Input**: The dataset of 2253 tweets was split into 80% and 20% training and validation datasets denoted as St and Sv respectively. Similarly, emoji only dataset of 855 records was also split into 80% and 20% training and test datasets.
2: **Output**: Sentiment(Positive/Negative)
3: **Begin**:
4: The raw datasets are then passed into the S-BERT and Universal Sentence Encoder models to generate embeddings. Where embeddings of fixed length 1024 and 512 respectively are generated for all the train and test datasets.
5: *for* (train,test) in onlytextdataset(train,test), onlyemojidataset(train,test),
both(train,test) *do*
6: Et1=S-BERT(train) Train embeddings
7: Ev1=S-BERT(test) Test embeddings
8: Et2=USE(train) Train embeddings
9: Ev2=USE(test) Test embeddings .
10: end for
11: The train embeddings are used to train the Standard fully connected NN and LSTM NN models.

12: **for** (t,v) in (Et1,Ev1),(Et2,Ev2) **do**
13: model1= Standard NN(t)
14: model2= LSTM NN(t)
15: end **for**
16: Finally, validation or test datasets are used to validate the model results.
17: accuracy.collect(model1.evaluate(t, v))
18: accuracy.collect(model2.evaluate(t,v))

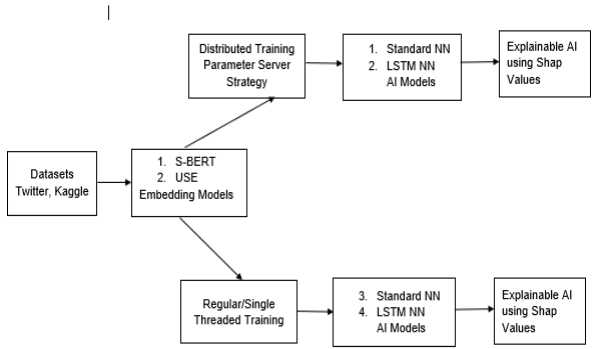

Figure 1. Overall methodology.

*A. Embedding Models: SBERT and USE*

We built two models, using the Sentence Bidirectional Encoder Representations from Transformers (SBERT) and Universal Sentence Encoder (USE). for generating sentence embeddings. Context-free models such as word2vec or GloVe generate a single word embedding representation for each word in the vocabulary. Contextual models instead generate a representation of each word that is based on the other words in the sentence. BERT, as a contextual model, captures these relationships in a bidirectional way. Sentence-BERT (SBERT), a modification of the pre-trained BERT network uses siamese and triplet network structures to derive semantically meaningful sentence embeddings that can be compared using cosine-similarity. The siamese network architecture enables that fixed-sized vectors for input sentences can be derived. Using a similarity measure like cosine-similarity or Manhatten / Euclidean distance, semantically similar sentences can be found. These similarity measures can be performed extremely efficient on modern hardware, allowing SBERT to be used for semantic similarity search as well as for clustering. For more information on SBERT one can refer to [14]. The Universal Sentence Encoder model encodes textual data into high dimensional vectors known as embeddings which are numerical representations of the textual data. It specifically targets transfer learning to other NLP tasks, such as text classification, semantic similarity, and clustering. The pre-trained Universal Sentence Encoder is publicly available in Tensorflow-hub. It comes with two variations i.e. one trained with Transformer encoder and the other trained with Deep Averaging Network (DAN). The two have a trade-off of accuracy and computational resource requirement. While the one with Transformer encoder has higher accuracy, it is computationally more intensive. The one with DNA encoding is computationally less expensive and with little lower accuracy. For more information on USE one can refer [5].

*B. Classification Models: Standard Fully Connected and LSTM Neural Networks*

Long Short Term Memory (LSTM) networks are a type of Recurrent Neural Networks (RNN). They can learn order dependence in sequence prediction problems.

An artificial neural network is a layered structure of connected neurons, inspired by biological neural networks. RNN is a class of neural networks tailored to deal with temporal data. The neurons of RNN have a cell state/memory, and input is processed according to this internal state, which is achieved with the help of loops within the neural network. At a high level, a Recurrent Neural Network (RNN) processes sequences like daily stock prices, sentences, or sensor measurements one element at a time while retaining the memory (called a state) of what has come previously in the sequence.

At the heart of an RNN is a layer made of memory cells. The most popular cell at the moment is the Long Short-Term Memory (LSTM) which maintains a cell state as well as a carry for ensuring that the signal (information in the form of a gradient) is not lost as the sequence is processed. At each step, the LSTM considers the current word, the carry, and the cell state. Anatomy of LSTM can be seen in Fig. 2. The LSTM has 3 different gates and weight vectors: there is a forget gate for discarding irrelevant information; an input gate for handling the current input, and an output gate for producing predictions at each time step. For more information on LSTM one can refer to The Unreasonable Effectiveness of Recurrent Neural Networks [15].

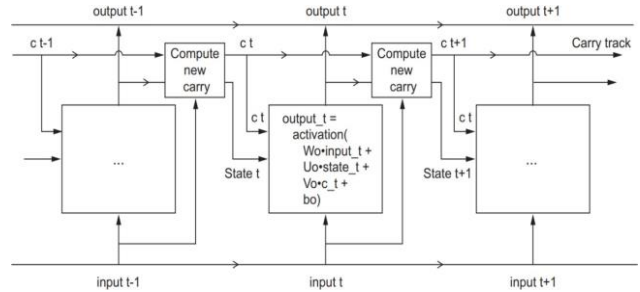

Figure 2. Anatomy of LSTM.

*C. Distributed Training*

Once we built the sentence embeddings we applied the transfer learning technique to integrate the embeddings with the Keras Sequential DNN model. Besides, in general, models are trained on a single machine; however, in real-time, the models have to be trained on billions of data points. Therefore, models will take a lot of time to complete, to overcome this problem distributed training has been introduced [16]. We have applied the Parameter server strategy to train the models on regular AWS EMR clusters without any GPUs. The Parameter server strategy is a data-parallel method to scale up model training on multiple machines. A parameter server strategy cluster consists of workers and parameter servers. Variables are created on parameter servers and they are read and updated by workers in each step. By default, workers read and update these variables independently without synchronizing with each other. For more information on Parameter server strategy, one can refer [17].

*D. Explainable AI*

Once models are built we would like to know if there are any biases or discrimination against users. To validate these points and make sure the model is blind to discrimination we have used explainable AI (XAI). Explainable AI (XAI) [18] methods are used to explain the learning and decisions predicted by a neural network. Using XAI it is now possible to learn where and how a Neural Network needs improvement and what are the strengths and weaknesses of the model that was trained. Shap algorithm is the most commonly used approach for examining the model biases. SHAP (SHapley Additive exPlanations) is a game-theoretic approach to explain the output of any machine learning model. It connects optimal credit allocation with local explanations using the classic Shapley values from game theory and their related extensions. SHAP assigns each feature an importance value for a particular prediction. Its novel components include: (1) the identification of a new class of additive feature importance measures, and (2) theoretical results showing there is a unique solution in this class with a set of desirable properties. For more information on SHAP one can refer [19].

## V. EXPERIMENTS

The H4EAD data was collected using the tweeter search API over a period of one week. The positive tweets were collected by querying the "saveh4ead" hashtag and the negative tweets were collected by querying the "noh4ead" hashtag. There were 2252 tweets, out of which 246 were negative and 2006 were positive tweets. Since tweets are sentences we have used sentence embedding models like Universal Sentence Encoder and SBERT to embed sentences to sentence vectors instead of the basic Bag of Words (BOW) models like TF-IDF and Count vectors. In addition, sentences can also contain emojis and a few users may respond using emojis only, to address this problem we have used Kaggle [20] emoji dataset that rank the emojis as positive and negative. Based on this we selected a few positive and negative sentiment emojis.

We ran the following experiments 1. Classifying only text 2. Classifying only Emojis, and finally 3. Classifying tweets containing both. The experiments were repeated using each model. Table I and Table II show results of running S-BERT embeddings with Standard and LSTM NN. Similarly, Table III and Table IV show the results of running Universal Sentence Encoder embeddings with Standard and LSTM NN respectively.

TABLE I. S-BERT STANDARD NN

| Dataset | Precision | Recall | F-Score | Accuracy |
|---|---|---|---|---|
| Text Only | 0.98 | 0.992 | 0.985 | 0.986 |
| Emoji Only | 0.70 | 0.72 | 0.71 | 0.710 |
| Text and Emoji | 0.997 | 0.995 | 0.995 | 0.996 |

TABLE II. S-BERT LSTM

| Dataset | Precision | Recall | F-Score | Accuracy |
|---|---|---|---|---|
| Text Only | 0.978 | 0.968 | 0.973 | 0.973 |
| Emoji Only | 0.711 | 0.73 | 0.720 | 0.720 |
| Text and Emoji | 0.98 | 0.97 | 0.975 | 0.975 |

TABLE III. UNIVERSAL SENTENCE ENCODER STANDARD NN

| Dataset | Precision | Recall | F-Score | Accuracy |
|---|---|---|---|---|
| Text Only | 0.987 | 0.985 | 0.996 | 0.986 |
| Emoji Only | 0.723 | 0.720 | 0.7249 | 0.725 |
| Text and Emoji | 0.975 | 0.975 | 0.975 | 0.975 |

TABLE IV. UNIVERSAL SENTENCE ENCODER (USE) LSTM

| Dataset | Precision | Recall | F-Score | Accuracy |
|---|---|---|---|---|
| Text Only | 0.99 | 1 | 0.995 | 0.995 |
| Emoji Only | 0.722 | 0.721 | 0.721 | 0.7215 |
| Text and Emoji | 0.996 | 0.992 | 0.994 | 0.994 |

*A. Distributed Training*

The parameter server strategy was used for conducting distributed training with 5 threads. The runtime improvements are shown in Table V.

*B. Explainable Artificial Intelligence*

Once models are built we would like to know if there are any biases or discrimination of users. To validate these points and make sure the model is blind to discrimination we have used explainable AI (XAI). Shap algorithm is the most commonly used approach for examining the model biases, we examined the Shap values for multiple tweets a sample of few tweets both text and emoji are shown in Fig. 3.

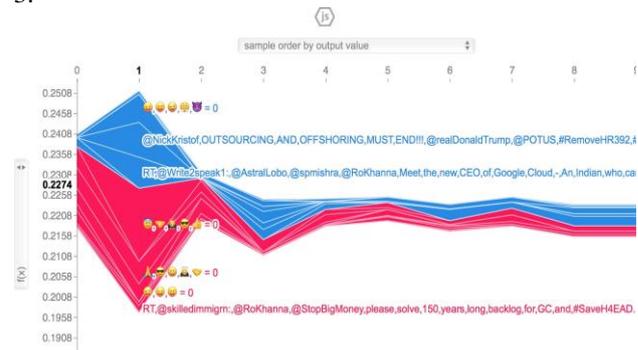

Figure 3. Shap Algorithm: Explainable AI.

## VI. RESULTS & ANALYSIS

From the experiments conducted, several observations are inferred. The text classification accuracy was almost the same for both the models around 98% accuracy, this is clearly an improvement over the 78% accuracy achieved in the paper [6]. However, when it comes to emojis the models identified the seen emojis with 100% accuracy, on the contrary, when the validation set was built using emojis that were not present in the training set then the accuracy of both the models reduced drastically to 70% by wrongly classifying all the unseen emojis. This is mainly because, as a word can have many similar words the same sentence can be worded in different ways and since we used sentence embedding models that are good at capturing the semantic similarities the models performed well when it saw completely new sentences. On the other hand, when it comes to emojis the models can identify the learned emojis with 100% accuracy but perform poorly when tested on new emojis. This shows that, both the Universal and S-BERT Sentence Embedding models are good for text embeddings; however, perform poor in identifying the semantic relationship between the emojis.

As we can see from Table V, distributed training with 5 threads took only 5.88 seconds consistently across multiple runs. Besides, the accuracy of the models remained the same in both standalone and distributed training methods. From Fig. 3 we have 7 tweets out of which 4 are emoji only and 3 are text only tweets. The red color shows the positive sentiments and blue color shows the negative sentiments, as we can see that positive and negative tweets are correctly classified based on the feature set. Besides, in our examination we didn't see any kind of model biases for the given feature set.

TABLE V. EXECUTION TIME IN SECONDS

| Mode of Training | Jupyter Notebook | Terminal |
|---|---|---|
| Single Thread | 47.4980 | 34.5948 |
| Distributed Training with 5 Threads | NA | 5.8868 |

VII. CONCLUSION AND FUTURE WORK

In this work, we performed sentiment analysis of Tweets as well as on emoji dataset from Kaggle. Since tweets are sentences we have used Universal Sentence Encoder and SBERT sentence embedding models to generate the embeddings. The embeddings are used to train the Standard fully connected NN, and LSTM NN models. We observed that the text classification accuracy was almost the same for both the models around 98%. On the contrary, when the validation set was built using emojis that were not present in the training set then the accuracy of both the models reduced drastically to 70%. In addition, the models were also trained using the distributed training approach instead of a traditional single-threaded model for better scalability. Using the distributed training approach, we were able to reduce the runtime by roughly 15% without compromising on accuracy. Finally, as part of explainable AI the Shap algorithm was used to explain the model behaviour and check for model biases for the given featureset. As part of future direction, this work could be extended to multilingual datasets and we are also planning to collect millions of data points and evaluate the model performance on large datasets.

CONFLICT OF INTEREST

The authors declare no conflict of interest.

AUTHOR CONTRIBUTIONS

Dr. Sirisha Velampalli: Research Idea selection, Literature review, writing the paper, and guiding the research. Read existing papers and identify the research question, and define the scope. Mr. Chandrashekar Muniyappa: Design, Develop, and Experiment the research idea. Identify the right tech stack and implement the required logic to get the idea working. Dr. Ashutosh Saxena: Final Review and Feedback on the paper.

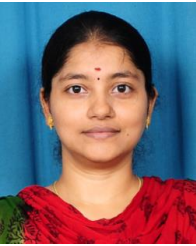

**Dr. Sirisha Velampalli** is currently working as Assistant Professor in CR Rao AIMSCS, University of Hyderabad Campus, Hyderabad. Her research interests are in the areas of Graph Mining, Graph Based Anomaly Detection, Big Data Analytics. She obtained Ph.D. under TEQIP-II funding from University College of Engineering, JNTU Kakinada. She has made research presentations at national and international conferences, workshops and has published articles in Data Mining, Big Data Analytics, Graph Based Anomaly Detection. She teaches computer science courses for graduate and undergraduate programs.

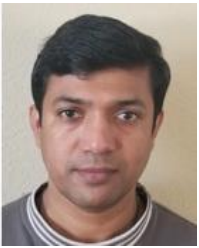

**Chandrashekar Muniyappa** received Master's in Data Science from the University of Wisconsin Green Bay. He has around 14 years of experience in Software development and has worked at Yahoo, AOL in the past and currently is employed at Samsung Electronics. He has been awarded with multiple patents and IEEE publications. Data mining, statistics, big data, Operational Research, Data structurers and Algorithms are his research areas.

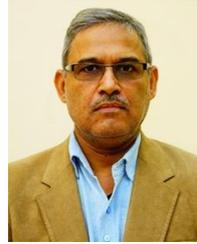

**Dr. Ashutosh Saxena** is an industry expert and academician with over two decades of experience, 100+ international publications with h-index of 16, along with 31 filled patents out of which 27 are now US granted patents. Two books, one on "PKI: Concept, Design and Deployment" (Tata McGraw Hill Publications 2004) and second on "Blockchain Technology: Concepts and Applications", (Wiley Publications 2020) to his credit. Research interest is in the areas of information security and privacy. Earned a Master of Science (1990), Master of Technology (1992), and a Doctorate (PhD) in computer science (1999) from Devi Ahilya University, Indore, and has qualified CSIR-UGC examination. Also received the BOYSCAST Post- Doctoral Fellowship (2001-2002) at Information Security Research Center, Queensland University of Technology, Brisbane, Queensland, Australia. Successfully completed Certification on Intellectual Property (2007) organized by World Intellectual Property Organization and Post-Graduation Diploma in Patents Law (2009) from NALSAR, University of Law, Hyderabad. Membership of Professional Associations:
• Life Member of Computer Society of India;
• Life Member of Cryptology Research Society of India;
• Senior Member of IEEE Computer Society.